\documentclass[10pt,twocolumn,letterpaper]{article}

\usepackage{cvpr}
\usepackage{times}
\usepackage{epsfig}
\usepackage{graphicx}
\usepackage{amsmath}
\usepackage{amssymb}
\usepackage[skip=1pt]{caption}
\usepackage{subcaption}
\usepackage{float}
\usepackage{multirow}
\usepackage{xcolor}

\usepackage[pagebackref=true,breaklinks=true,letterpaper=true,colorlinks,bookmarks=false]{hyperref}

\cvprfinalcopy 

\def\cvprPaperID{5337} 

\ifcvprfinal\fi
\begin{document}

\newcommand{\STAB}[1]{\begin{tabular}{@{}c@{}}#1\end{tabular}}

\title{Multiple Instance Captioning: Learning Representations from Histopathology Textbooks and Articles}

\author{Jevgenij Gamper\\
Warwick University \\
{\tt\small j.gamper@warwick.ac.uk}
\and
Nasir Rajpoot\\
Warwick University\\
{n.m.rajpoot@warwick.ac.uk}
}

\twocolumn[{%
\renewcommand\twocolumn[1][]{#1}%
\maketitle
\begin{center}
    \centering
    \includegraphics[width=\linewidth]{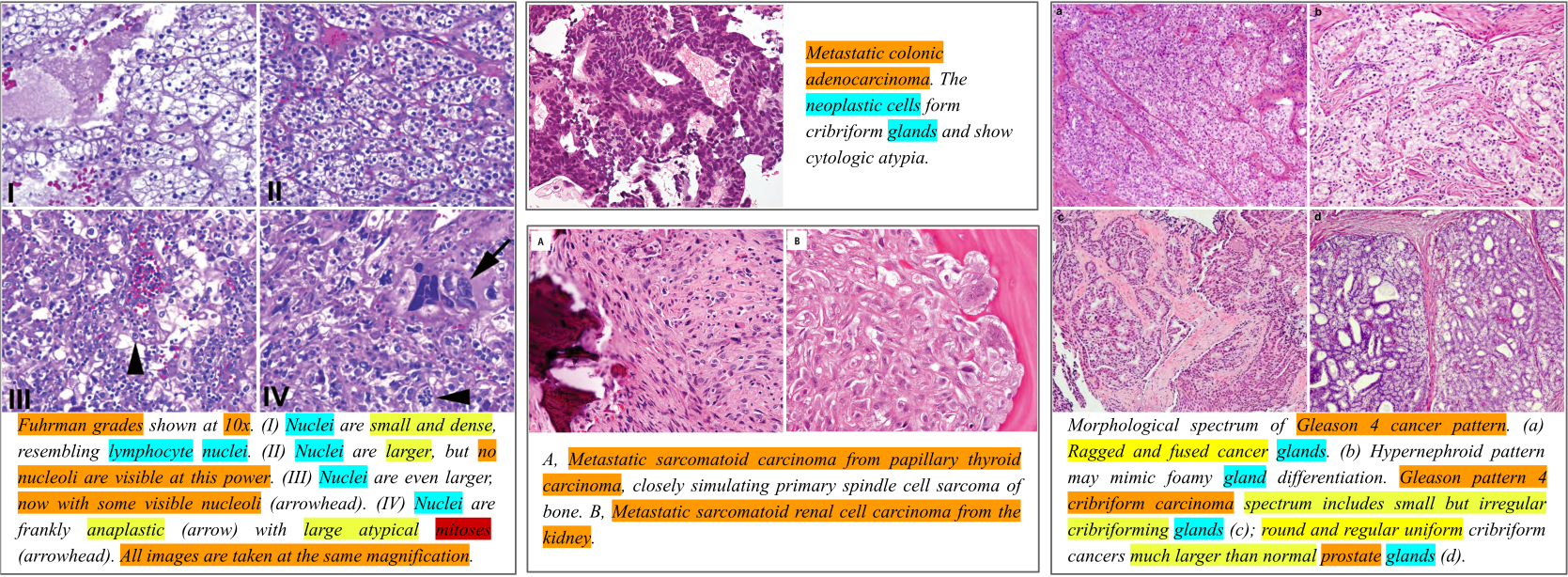}
    \captionof{figure}{Four samples from ARCH, a multiple instance captioning computational pathology dataset. Samples on the left and right each consist of four image instances with a single caption; top-middle shows an image-caption pair while bottom-middle contains two image instances with a single caption. Labeled in color are examples of common tasks within computational pathology: diagnostic (orange); detection \& classification (cyan); descriptive (yellow); special cell detection (red).}
    \label{fig:caption_examples}
\end{center}%
}]

\begin{abstract}
   We present ARCH, a computational pathology (CP) multiple instance captioning dataset to facilitate dense supervision of CP tasks. Existing CP datasets focus on narrow tasks; ARCH on the other hand contains dense diagnostic and morphological descriptions for a range of stains, tissue types and pathologies. Using intrinsic dimensionality estimation, we show that ARCH is the only CP dataset to (ARCH-)rival its computer vision analog MS-COCO Captions. We conjecture that an encoder pre-trained on dense image captions learns transferable representations for most CP tasks. We support the conjecture with evidence that ARCH representation transfers to a variety of pathology sub-tasks better than ImageNet features or representations obtained via self-supervised or multi-task learning on pathology images alone. We release our best model and invite other researchers to test it on their CP tasks.
\end{abstract}

\section{Introduction}

The success of an intelligent system depends on having the right representation of data. Computer vision community has gradually moved from engineering features to letting the deep neural networks learn data representations, given a differentiable task objective \cite{alom2018history, garcia2017review}. 

\begin{figure*}
\begin{center}
\includegraphics[width=\linewidth]{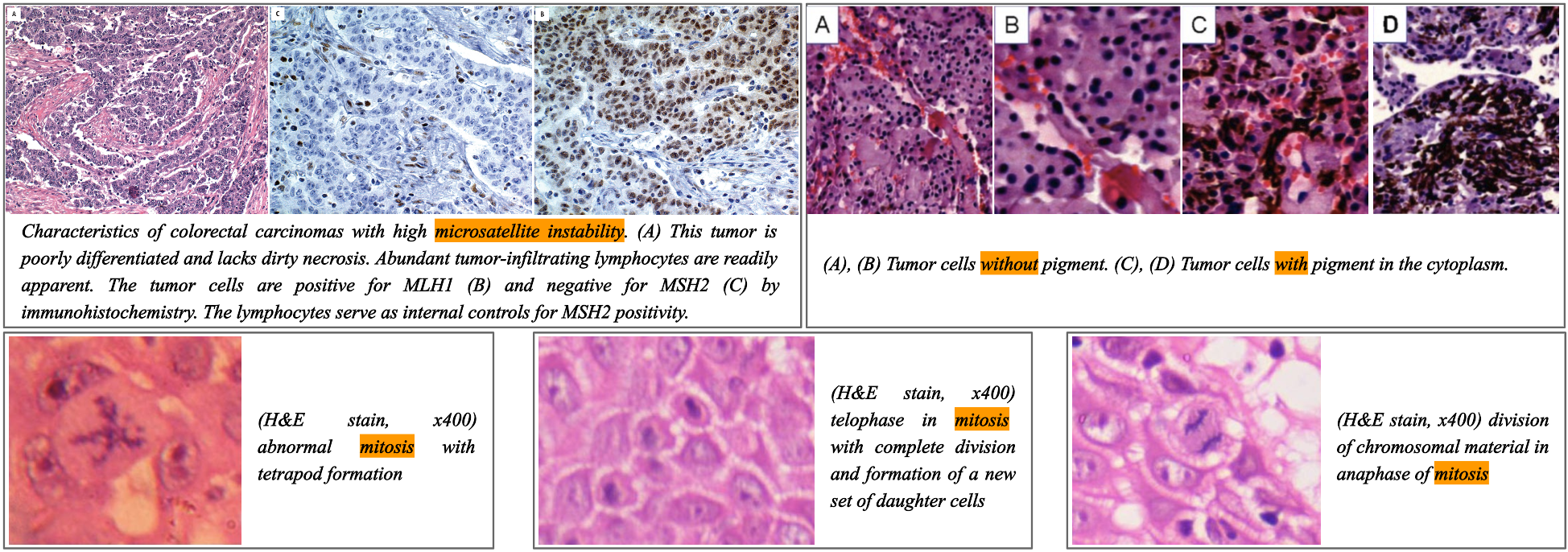}
\end{center}
   \caption{Five samples from ARCH. {\em Top left}: Caption describes morphological characteristics of Microsatellite Instability (MSI) in colon cancer; {\em Top right}: Caption provides contrastive supervision where cells in first two images have to be contrasted against pigmented cells in the last two images;
   {\em Bottom row}: Captions provide morphological description of cell-division or {\em mitosis} at different stages.}
\label{fig:second_title_fig}
\end{figure*}

Computational pathology (CP), a sub-field of medical imaging that entails quantitative profiling of spatial patterns in multi-gigapixel whole-slide images (WSIs) of patient tissue slides, has followed these developments closely \cite{colling2019artificial}. Deep Learning (DL) has been applied to detecting cancerous regions \cite{kather2016multi}, classifying tissue sub-types \cite{javed2020cellular}, identifying diagnostically relevant structures such as glands \cite{graham2019mild}, nuclei \cite{gamper2019pannuke}, vessels and nerves \cite{fraz2019fabnet}, quantifying spatial patterns of tumor infiltrating lymphocytes \cite{saltz2018spatial} and histology image retrieval \cite{hegde2019similar}. More recently, DL been used to learn representations for challenging tasks of predicting genetic sub-types \cite{fu2020pan, kather2020pan}.

With recent advances in weak and unsupervised learning, it is becoming widely acceptable to design computer vision systems from pre-trained representations \cite{tung2017self, godard2019digging, hendrycks2019using, jing2020self, liu2019selflow, goyal2019scaling}. CP community has followed suit, by relying less on patch-level annotations and instead adopting weakly supervised methods such as multiple instance learning (MIL) \cite{campanella2019clinical, lu2020data, ilse2018attention} or neural image compression (NIC) \cite{tellez2019neural} to perform inference directly from WSI-level data. However, most of the above methods rely on encoders pre-trained on ImageNet for feature extraction.  

These developments create a growing demand for pre-trained pathology image representations. In this paper, we present ARCH -- a new dataset of dense image captions mined from clinical and academic pathology textbooks and articles -- and demonstrate the universality of pathology image features obtained by pre-training on ARCH. 

We conjecture that \textit{the global minimum of the multiple instance captioning objective on a dataset like ARCH is the global minimum of the majority of computational pathology sub-tasks commonly solved and published to date}.
\begin{figure*}
\begin{center}
\includegraphics[width=\linewidth]{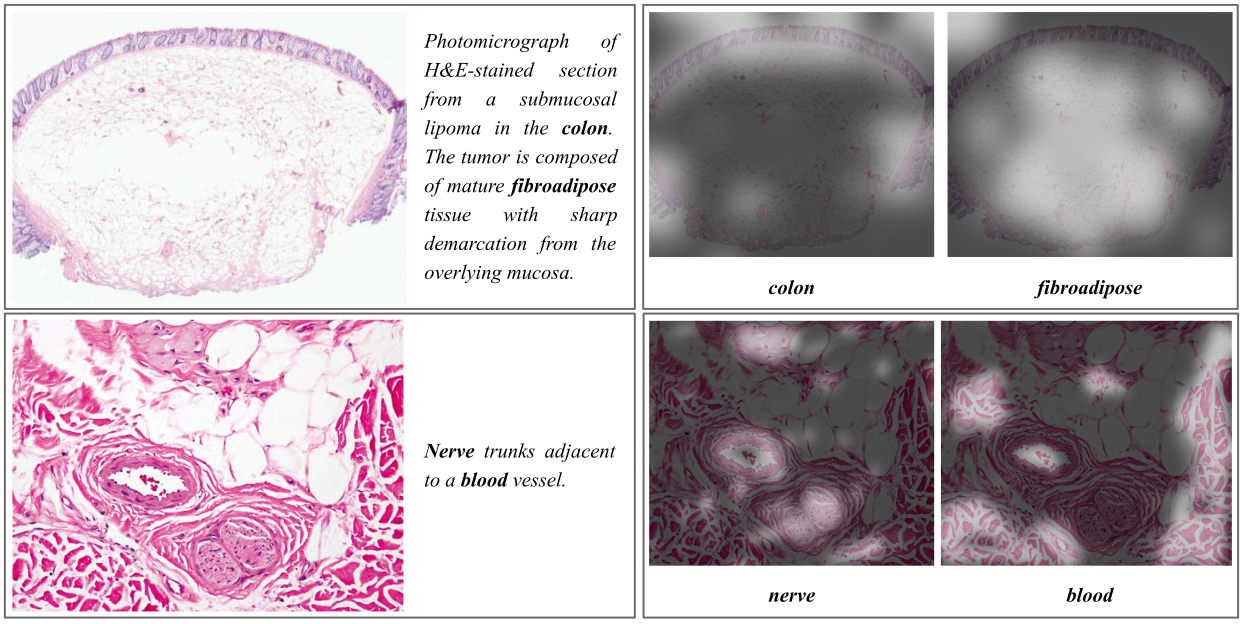}
\end{center}
   \caption{{\em Left}: samples of tissue at low and higher power view with their corresponding captions; {\em Right}: Model generated attention masks the corresponding words highlighted in bold. Attention masks were generated according to the methods described in Desai \etal\cite{desai2020virtex}, see Section \ref{sec:results} for details.}
\label{fig:attention_plot}
\end{figure*}
We base this conjecture on several observations made while putting together this dataset and on experimental results reported in this work. In Figure \ref{fig:caption_examples}, correctly captioning a bag of four image instances (left) requires an algorithm to: identify nuclei as well as their category (cyan) \cite{gamper2019pannuke}, describe nuclear characteristics (yellow) \cite{lee2017nuclear}, detect mitotic figures (red) \cite{cirecsan2013mitosis, veta2019predicting} and even understand the relationship between image magnification and diagnosis (orange). Besides identifying cells and glands (all examples, cyan) \cite{fraz2019fabnet, graham2019mild}, an algorithm has to learn to describe an image as being metastatic tumour originating in colon (middle-top) or a metastatic sarcoma from papillary thyroid (middle-bottom). Only recently, Lu \etal\cite{lu2020deep} proposed a method for addressing the challenging task of identifying metastatic cancers and tumor origins. In Figure \ref{fig:caption_examples} on the right, an algorithm has to identify the bag of four images as prostate and assign it a Gleason grade 4 \cite{nagpal2019development, arvaniti2018automated, bulten2020automated}. The ARCH multiple instance captioning dataset stands out by offering implicit supervision of contrastive learning -- where instances within the same bag have to be contrasted in order to generate the caption (Figure \ref{fig:second_title_fig} top-right). ARCH includes samples that provide explicit supervision on genetic characteristics expressed through tissue morphology (Figure \ref{fig:second_title_fig} top-left), a task that has only recently been explored \cite{kather2019deep, kather2020pan, fu2020pan}. One may also find close up descriptions of mitosis at various stages (Figure \ref{fig:second_title_fig} bottom row) \cite{roux2014mitos}. We provide additional samples from ARCH in the Supplementary Material.
%
%
We also present attention plots (see Figure \ref{fig:attention_plot}) as additional evidence to support our conjecture and demonstrate examples at low and medium resolutions that are widely available in ARCH.

While ARCH includes unique CP tasks as sub-tasks to image captioning, we believe it is the relationship between the sub-tasks described in captions that provides unique supervision during pre-training and results in more general CP image features. In Section \ref{sec:intrinsic-dim}, we compare the intrinsic dimensionality of CP datasets to their analogs in computer vision. CP datasets appear to have lower intrinsic dimensionality than CIFAR100 and CityScapes in their respective tasks of image classification and segmentation. ARCH, on the other hand, matches the complexity of its computer vision counterpart, the MS-COCO Captions dataset. Recently, Desai \etal\cite{desai2020virtex} showed that pre-training on MS-COCO captions is superior to ImageNet. This might provide a hint to why ARCH pre-training is advantageous to multi-task training. Individual CP tasks provide narrow supervision and even when combined together in a multi-task setting, the supervision signal does not explicitly model the relationship between the tasks.

Finally, besides offering a new computer vision task of multiple instance captioning, ARCH presents a previously unexplored path towards providing information rich ground truth for medical imaging, a field known for its annotation scarcity.

\begin{figure*}
     \centering
     \begin{subfigure}[b]{0.3\textwidth}
         \centering
         \includegraphics[width=\textwidth]{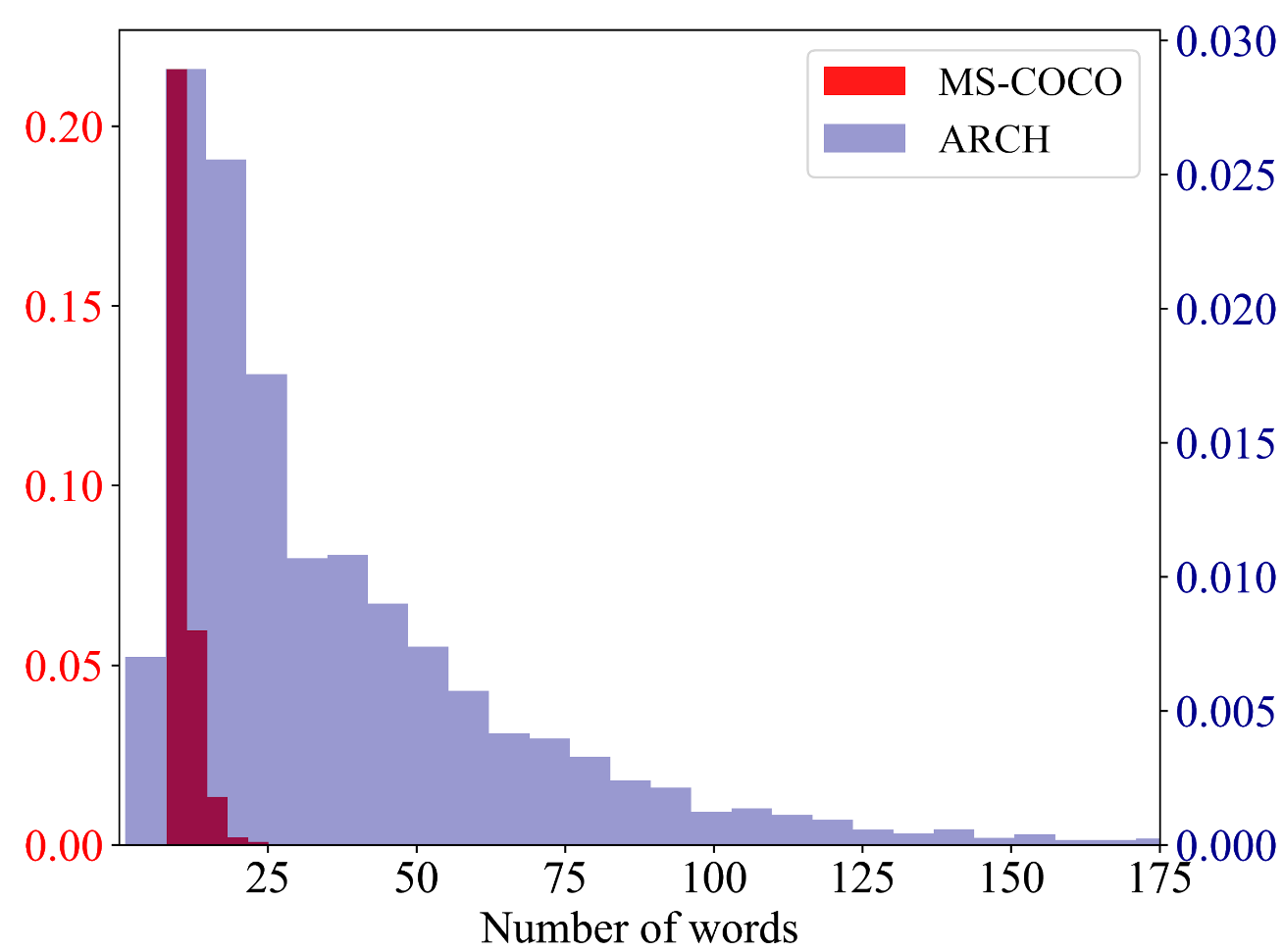}
         \caption{}
         \label{fig:caption_lengths}
     \end{subfigure}
     \hfill
     \begin{subfigure}[b]{0.3\textwidth}
         \centering
         \includegraphics[width=\textwidth]{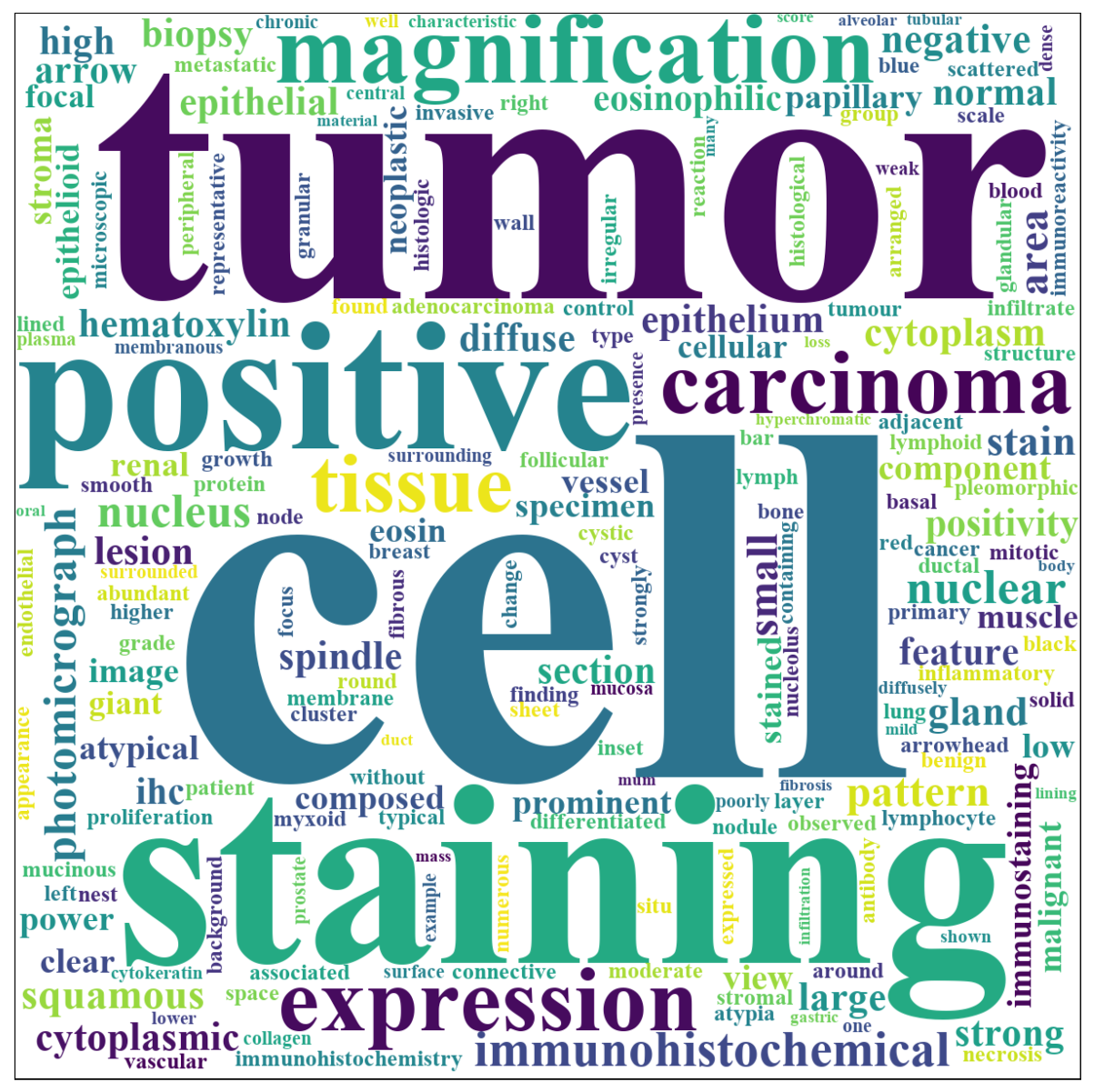}
         \caption{}
         \label{fig:word_cloud}
     \end{subfigure}
     \hfill
     \begin{subfigure}[b]{0.3\textwidth}
         \centering
        \begin{tabular}{lccc}
        \hline
        \multicolumn{1}{|c|}{Bag size}   & \multicolumn{1}{c|}{1}    & \multicolumn{1}{c|}{2}    & \multicolumn{1}{c|}{3}   \\ \hline
        \multicolumn{1}{|c|}{\# Bags} & \multicolumn{1}{c|}{9,772} & \multicolumn{1}{c|}{1,292} & \multicolumn{1}{c|}{461} \\ \hline
        \multicolumn{2}{r}{Total Bags}                               & \multicolumn{2}{l}{11,816}                             \\ \hline
        \multicolumn{1}{|l|}{Bag size}   & \multicolumn{1}{c|}{4}    & \multicolumn{1}{c|}{5}    & \multicolumn{1}{c|}{6}   \\ \hline
        \multicolumn{1}{|l|}{\# Bags} & \multicolumn{1}{c|}{173}  & \multicolumn{1}{c|}{44}   & \multicolumn{1}{c|}{36}  \\ \hline
        \multicolumn{2}{r}{Total Images}                             & \multicolumn{2}{l}{15,164}                            \\ \hline
        \multicolumn{1}{|l|}{Bag Size}   & \multicolumn{1}{c|}{7}    & \multicolumn{1}{c|}{8}    & \multicolumn{1}{c|}{9}   \\ \hline
        \multicolumn{1}{|l|}{\# Bags} & \multicolumn{1}{c|}{14}   & \multicolumn{1}{c|}{17}    & \multicolumn{1}{c|}{7}   \\ \hline
        \end{tabular}
         \caption{}
         \label{fig:five over x}
     \end{subfigure}
        \caption{(a) Histogram of caption lengths showing that MS COCO captions are significantly shorter as compared to captions in ARCH; (b) Word cloud of 1,000 most frequent words in ARCH captions; (c) Tables of bag sizes -- i.e., number of images per single caption in ARCH bags.}
        \label{fig:bag_sizes}
\end{figure*}

\section{Related Work}

Desai \etal\cite{desai2020virtex} showed that models pre-trained on MS-COCO Captions efficiently transfer to classification, and detection tasks. MS-COCO contains only 118k images, much smaller in number than ImageNet, yet the authors of \cite{desai2020virtex} have shown that learning from textual annotations is more data-efficient than from classification labels. They emphasized superior results from using more images with single captions, rather than multiple captions per image; this observation is particularly important for ARCH, where we only have a single caption per bag of images. In this work, besides offering a new dataset for multiple instance captioning in CP, we show that pre-training on single image captions is indeed more data-efficient and induces superior inductive biases for data representation as compared to multi-task or self supervised pre-training.

We use strict measures to compare pre-trained representations to support our conjecture. The standard methodology has been to extract features from the last layer of the trained encoder and evaluate them with a linear classifier \cite{ettinger-etal-2016-probing, subramanian2018learning, chen2020simple}, to which Resnick \etal\cite{resnick2019probing} referred to as \textit{linear probing} and showed it to be insufficient if downstream tasks are highly non-linear. In this work we benchmark the performance of an encoder pre-trained on ARCH against encoders pre-trained on ImageNet, via self- or multi-task supervision evaluated with the encoder that produces the best performance, while our model can only use linear encoder. 

Transferring models from ImageNet remained to be the dominant method in practice until recently, however pre-training on textual annotations and self-supervised learning are closing the performance gap. Kornblith \etal\cite{kornblith2018better} offered an empirical study on the transferability properties of models trained on ImageNet. 
Raghu \etal\cite{raghu2019transfusion} further studied ImageNet models transferability for medical imaging datasets and showed that the primary benefit of transfer has been for the initialisation of large model weights rather than feature transfer and that the same result could be obtained with far smaller models trained from scratch.

Meanwhile, the CP community has pursued two options to pre-train features that generalise across datasets and tasks to tackle the problem of the scarcity of ground-truth data: Multi-Task learning (MTL) and self-supervised (SS) learning \cite{lu2019semi}. Tellez \etal\cite{tellez2020extending} recently extended NIC via multi-task pre-training on four tasks. Mormont \etal\cite{mormont2020multi} provided an in-depth analysis of feature generalization of multi-task pre-training on 22 image classification tasks (11 of the used datasets are not publicly available). In this work, besides introducing a unique pre-training dataset ARCH, we use only publicly available datasets for evaluation.

Even though our goal is to obtain general image features from pre-training on image captions, not image captioning in itself, it is important to point out the work by Zhang \etal\cite{zhang2017mdnet}, who were the first to propose a pathology image captioning model and dataset. Yet the dataset used for training was small - 32 patient images that are not publicly available, limited to single tissue type (bladder) and a single stain type (H\&E). Our pathology image captions dataset is substantially larger, and includes a whole range of tissues, diagnoses and stain types. 

\begin{figure*}
\begin{center}
\includegraphics[width=\linewidth]{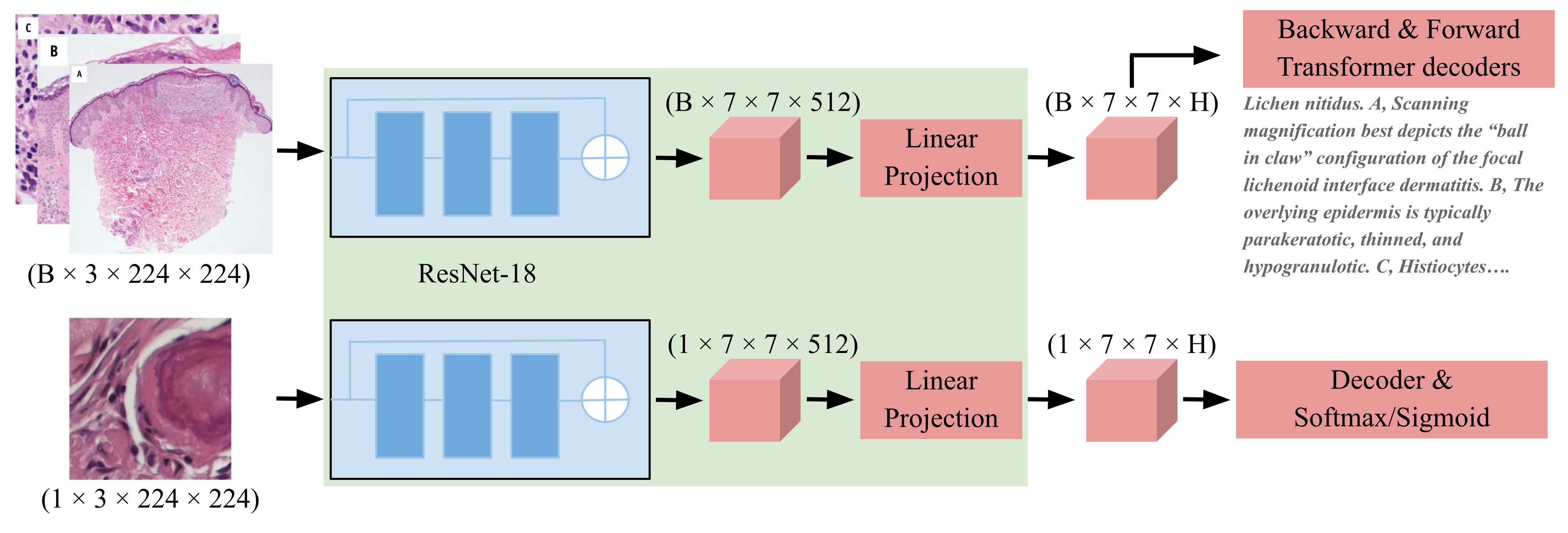}
\end{center}
   \caption{Schema of neural networks used in experiments, all use ResNet-18 encoder. {\em Top row}: ARCH pre-training architecture. Within a bag all images are encoded, these features are then projected by a linear layer with non-linearity and fed to backward and forward transformers as per Desai \etal\cite{desai2020virtex}; {\em Bottom row}: for image classification, after encoding the features are also projected by a linear layer with non-linearity and fed to sigmoid or softmax depending on the number of labels. In green, we show model components that are shared during multi-task training. For both models, we modify ResNet-18 to have the Batch-norm layer as an input layer.}
\label{fig:modeling}
\end{figure*}

\section{Method}

\subsection{ARCH Construction}

We used PubMed medical articles database and pathology textbooks to construct ARCH. 
We selected all PubMed journals according to keywords (\textit{pathology}, \textit{histochemistry}, \textit{histology}, \textit{histopathology}), which resulted in 12,676 journal articles on clinical and research pathology as of 2019. Using {\textit{pubmed parser}}\footnote{https://github.com/titipata/pubmed\_parser}, we extracted a total of 25,028 figures and their corresponding captions. We then manually selected 8,617 figure-caption pairs that contained at least one histology or imunohistochemistry (IHC) image and we saved text in caption that only related to the histology image. Individual images were then extracted from figures to create multiple instance bags with their respective captions that have been checked for formatting errors. When selecting images, we made sure that these did not include excessive text, marks and were of reasonable quality.

Using {\textit{PDF-Figures 2.0}}\footnote{https://github.com/allenai/pdffigures2}, we extracted figures and their captions from 10 textbooks, leading to 3,199 figure-captions pairs. We followed the above steps for constructing multiple instance bag and caption pairs, verifying the quality of the images as well as checking extracted captions for formatting errors.

ARCH contains 11,816 bags and 15,164 images in total. Figure \ref{fig:five over x} shows a more detailed breakdown by the number of bags according to the number of images within the bag, with the smallest bag size being 1 (9,772 samples) and the largest bag size being 9 for which we have only 7 samples.

\subsection{ARCH Intrinsic Dimensionality}
\label{sec:intrinsic-dim}

\begin{figure*}
\begin{center}
\includegraphics[width=\linewidth]{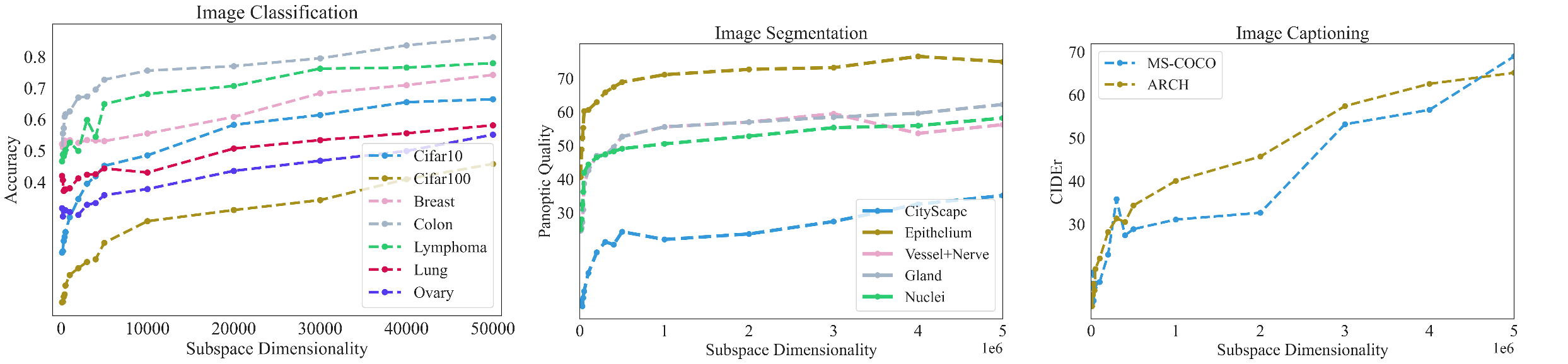}
\end{center}
   \caption{Relative complexity of computer vision and CP datasets for the tasks of image classification ({\em left}), segmentation ({\em middle}) and captioning (\em right).}
\label{fig:intrinsic_dimensionality}
\end{figure*}

Li \etal \cite{li2018measuring} demonstrated how a number of parameters in a randomly oriented subspace of a fixed neural network architecture can be used as a rough gauge of the difficulty of the problem. One slowly increases the dimension of this subspace, monitors at which dimension size the solutions first appear, and defines that to be the \textit{intrinsic dimensionality}. 
We follow this methodology to compare the relative complexity of CP datasets and their computer vision analogs, as shown in Figure  \ref{fig:intrinsic_dimensionality}.
\footnote{We used a publicly available PyTorch implementation of intrinsic dimensionality estimation \href{https://github.com/jgamper/intrinsic-dimensionality}{github.com/jgamper/intrinsic-dimensionality}}.
The three tasks are image classification, image segmentation and image captioning, with their respective evaluation metrics of accuracy, panoptic quality \cite{kirillov2019panoptic} and CIDEr \cite{vedantam2015cider} as per Desai \etal\cite{desai2020virtex}. For each of the tasks, as well as for all of our experiments in the paper, we used ResNet-18 \cite{he2015deep} as an encoder with its parameters wrapped in Fast-food transform \cite{le2013fastfood} for obtaining random sub-spaces required for intrinsic dimensionality estimation. All encoders were randomly initialised.

\subsection{ImageNet baseline} 

To benchmark representations we first need to obtain a competitive baseline. It is a general practice to use penultimate layer of a model trained on ImageNet for feature extraction, however some \cite{lu2020data} have found shallow layers to be more useful in CP tasks. In addition, \textit{linear probing} (LP), a method that uses a linear classifier to evaluate the quality of the representations may not be applicable. CP images have no reason to be linearly separable in ImageNet space. We thus search for optimal network depth for feature extraction i.e. the number of hidden layers in the decoder and dropout regularisation that optimally transfers ImageNet features for a given classification task. To achieve that, while keeping weights frozen of ResNet-18 pre-trained on ImageNet, we grid-searched for optimal decoder regularisation across 3 decoder depths (linear, 1, and 3 hidden layers) and 5 encoder depths. See Figure \ref{fig:detailed_results} for example results.

In Table \ref{tab:results}, a linear decoder on top of penultimate layer of ImageNet trained Resnet-18 is referred to as I-L; a linear decoder with an ImageNet ResNet-18 layer selected through hyper-parameter tunning is referred to as I-L-O; finally, I-B refers to the performance of the best ImageNet ResNet-18 layer followed by the decoder with depth and regularisation that lead to the best performance. 

\subsection{Self-supervised learning baseline} 
\label{sec:self-supervised}

Lu \etal\cite{lu2019semi} proposed contrastive predictive coding (CPC) as a self-supervised objective to pre-train a feature extractor before training a complete multiple instance learning model on whole slide images. We thus include this as an additional baseline for comparison labelled as SS. The training objective remains the same as in Lu \etal\cite{lu2019semi}, except that we use ResNet-18 instead of ResNet-50 as a feature encoder. 

\subsection{Multiple instance captioning} 
\label{sec:mic-section}

To pre-train a Multiple Instance Captioning (MIC) model on ARCH, we follow the setup of Desai \etal\cite{desai2020virtex}. We show the overall model scheme in Figure \ref{fig:modeling}-top, where the only difference with that of Desai \etal\cite{desai2020virtex} is that of working with bags of image instances instead of a single instance. We extend their code-base\footnote{Desai \etal\cite{desai2020virtex} code available at \href{https://github.com/kdexd/virtex}{github.com/kdexd/virtex}. Mormont \etal\cite{mormont2020multi} code used for multi-task learning available at \href{https://github.com/waliens/multitask-dipath}{github.com/waliens/multitask-dipath}.} to work in a multiple instance and/or multi-task setting. Given a dataset of bags of images with captions, our goal is to learn visual representations of images that can be transferred to downstream visual recognition tasks. 

A feature encoder ( ResNet-18 in our case) is used to process every image within a bag, leading to a representation of $\text{B} \times 7 \times 7 \times 512$ where $\text{B}$ is the size of the bag. Fully connected layer follows encoding features to $\text{B} \times 7 \times 7 \times \text{H}$, and fed into forward and backward transformers that perform  forward and backward captioning respectively. All of the model components are randomly initialized, and jointly trained to maximize the log-likelihood of the correct caption tokens $C = \{c_1, ..., c_{T} \}$:

\begin{align}
\mathcal{L}(\theta, \phi) = \sum_{t=1}^{T} \log (c_t | c_1, ..., c_{t-1}; \theta, \phi_f) \\ + \sum_{t=1}^{T} \log (c_t | c_{t+1}, ..., c_{T}; \theta, \phi_b)
\end{align}
where $\theta, \phi_f, \phi_b$ are the parameters of the feature extractor and linear layer, forward, and backward models respectively. Tokens are indexed by $t$.
Backward and forward transformer heads are used exclusively in order to backpropogate the supervision signal from the ground truth captions to the feature encoder - again, the goal here is not to learn the best image captioning model. 

\begin{figure}
\begin{center}
\includegraphics[width=\linewidth]{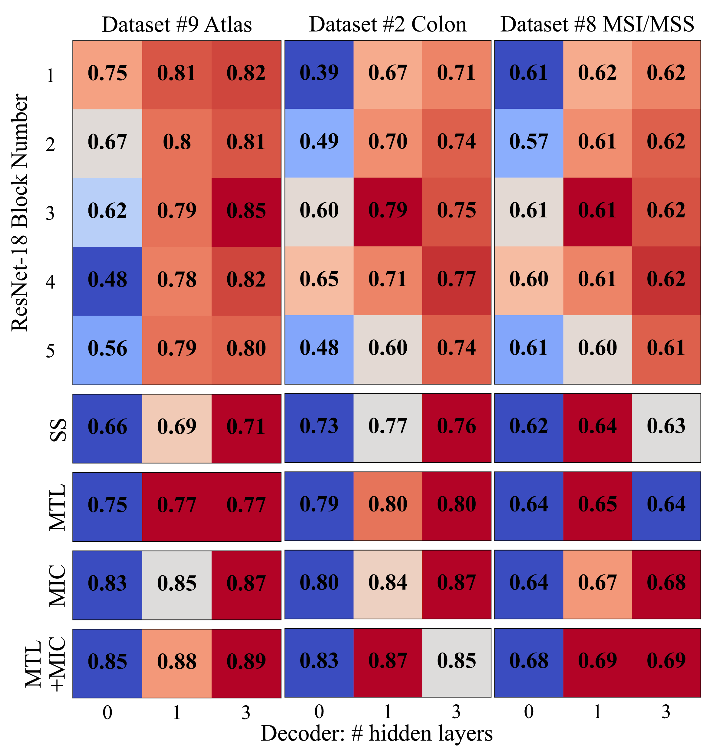}
\end{center}
   \caption{A study of pre-trained feature performance ($y$-axis) vs number of hidden layers in the decoder ($x$-axis). First five rows correspond to performance of features extracted from 5 residual blocks at a different depth in ResNet-18 trained on ImageNet. The remaining models correspond to: SS - self-supervised described in Section \ref{sec:self-supervised}; MTL - multi-task learning model described in Section \ref{sec:mtl-section}; MTL+MIC, a multi-task model trained along with ARCH dataset. All features are evaluated with a decoder with 0 (linear), 1 and 3 hidden layers with regularisation optimisation via grid-search. See Supplementary Material for all 12 datasets.}
\label{fig:detailed_results}
\end{figure}

The performance of the model trained on ARCH is referred to as MIC in Figures \ref{fig:intrinsic_dimensionality}, \ref{fig:detailed_results} and Table \ref{tab:results}. For all MIC model training experiments, we set the hyper-paremeters, tokenization and training details according to Desai \etal\cite{desai2020virtex} and their publicly available code, with a few exceptions as follows: H is set to 512, which also determines the width of each of the transformer layers and the number of attention heads; the batch size is set to 32 images or less irrespective of the bag sizes due to computational restraints, which are pre-computed before every epoch after re-shuffling the dataset indices. Finally, for the ease of training, we switched to a ADAM optimizer with a default learning rate of 1e-3 and an early stopping set to a patience of a held-out set validation loss of 10. 

\subsection{Multi-task learning} 
\label{sec:mtl-section}

We also experiment with adding MIC as part of the overall multi-task training with other datasets at hand in the same vein as Mormont \etal\cite{mormont2020multi}, the idea being that linear decoders may enforce linear separability of the learnt features.

In the MTL setup, the ResNet-18 encoder is shared among the transformer decoders for learning from the caption data and linear decoders for the 12 tasks specified in Table \ref{tab:results}. We refer to this model as MTL+MIC, however we also report results of multi-task training without MIC objective as an additional baseline. When training the MTL+MIC model, we had to resort to gradient accumulation to be able to use all datasets at every training iteration.

For all of our MIC, MTL and MTL+MIC models, we employ standard data augmentations using \href{https://imgaug.readthedocs.io/en/latest/}{imaug}: custom random crop function such that letters are not cropped out); resize and re-scale; color jitter (brightness, contrast saturation and hue); Gaussian and/or salt and pepper noise, and JPEG compression artifacts. 

\begin{table*}[]
\centering
\begin{tabular}{|c|c|c|c|c|c|c|c|c|c|}
\hline
Num. & Dataset (\# classes) & Tissue/Task & I-L  & I-L-O & I-B  & SS  & MTL & MIC & MTL+MIC \\ \cline{1-10}
1 & Litjens \etal\cite{litjens20181399} (2) & Breast & 50.9 & 66.1 & 72.3 & 73.2 & \textbf{91.8} & 90.5 & 90.0  \\ 
2 & Kather \etal\cite{kather2019predicting} (9) & Colon & 60.6 & 65.7 & 79.9  &  77.4  & 80.5 & 81.7 & \textbf{83.9}  \\
3 & Alsubaie \etal\cite{alsubaie2018multi} (6) & Lung & 43.4 & 52.0  & 60.4 & 73.4 & 78.6 & 82.4 & \textbf{87.0} \\ 
4 & Janowczyk \etal\cite{janowczyk2016deep} (3) & Lymphoma & 18.6 & 23.9 & 26.4 & 45.2 & 44.2 & 42.1 & \textbf{47.3} \\
5 & Qureshi \etal\cite{qureshi2008adaptive} (4) & Meningioma & 20.6 & 36.7 & 51.1 & 56.6 & 70.0 & 73.9 & \textbf{84.8} \\
6 & Shaban \etal\cite{shaban2019novel} (3) & Head \& Neck & 33.0 & 39.1 & 66.0 & 68.3 & 70.9 & 73.2 & \textbf{75.2} \\
7 & Kobel \etal\cite{kobel2010diagnosis} (5) & Ovary & 22.1 & 31.4 & 67.6 & 65.1 & 72.5 & 74.7 & \textbf{77.0} \\
8 & Kather \etal\cite{kather2019deep} (2) & MSI/MSS & 61.4 & 61.4 & 62.3 & 64.5 & 64.0 & 64.5 & \textbf{68.1} \\ 
9 & Hosseini \etal\cite{hosseini2019atlas} (22) & Multi-label & 56.9 & 75.0 & 85.6 & 77.5 & 75.8 & 83.7 & \textbf{85.4} \\ 
10 & Arvaniti \etal\cite{arvaniti2018automated} (5) & Gleason Scoring & 18.1 & 35.3 & 61.5 & 68.1 & 67.2 & 75.2 & \textbf{79.4} \\ 
11 & Veta \etal\cite{veta2019predicting}+\cite{janowczyk2016deep, roux2014mitos} (2) & Nuclear Mitosis & 58.7 & 59.0 & 63.4 & 57.5 & 68.5 & \textbf{70.4} & 69.3 \\ 
12 & Roux \etal\cite{roux2014mitos} (2) & Nuclear Atypia & 49.3 & 52.6 & 67.3 & 66.1 & 69.4 & 75.1 & \textbf{80.2} \\ \cline{1-10}
\end{tabular}
\caption{Accuracies for the classification datasets obtained with: I-L, penultimate ImageNet trained ResNet-18 layer and linear decoder; I-L-O best ImageNet trained ResNet-18 layer and linear decoder; I-B  best ImageNet trained ResNet-18 layer and optimal decoder depth; SS, a self-supervised learning model described in Section \ref{sec:self-supervised}; MTL, penultimate multi-task trained ResNet-18 layer and linear decoder; MIC, penultimate ARCH trained ResNet-18 layer and linear decoder; MTL+MIC, penultimate ARCH and multi-task trained ResNet-18 layer and linear decoder. Note that dataset \#9 is a multi-label classification problem, for which we report average accuracy per binary label to match the original work.}
\label{tab:results}
\end{table*}

\section{Experimental Results}
\label{sec:experiments}

The goal of our experiments section is to show that pre-training a ResNet-18 encoder on ARCH captions allows the encoder to learn transferable features; it is not our goal to learn an image captioning model. As such, we aim to demonstrate the potential of yet unexplored source of supervision -- medical textbooks and articles, and their combination with publicly available labelled data. 

\subsection{Setup}
To obtain results in Table \ref{tab:results} and Figure \ref{fig:detailed_results} for SS, MTL and MTL+MIC models, we follow the experimental setup described in Mormont \etal\cite{mormont2020multi} -- i.e., leaving one dataset out. For instance, to test the performance of SS model on the breast cancer classification dataset (Table \ref{tab:results} dataset \#1), we exclude that dataset from training and pre-train the SS model on the remaining 11 datasets. The same would apply to MTL and MTL+MIC models, except that we would never exclude ARCH from training in the latter. For ImageNet baselines, we only train on the specific dataset that we are interested in obtaining performance for. 

\subsection{Results and Discussion}
\label{sec:results}

As expected, features for CP images derived from ResNet-18 pre-trained on ImageNet are not linearly separable, as shown by the result for I-L column in Table \ref{tab:results}. It is worth noting, however, that there is a considerable amount of variability in the performance of ImageNet features depending on the ResNet-18 block chosen for feature extraction and the number of hidden layers in the decoder. For example, for the colon tissue sub-typing (dataset 2, Table \ref{tab:results}), the accuracy can range from 39\%  to 79\% (see Figure \ref{fig:detailed_results}), and yet remain almost the same across all combinations for the task like MSI/MSS classification. 
In our experiments, we did not see a clear pattern for an optimal decoder depth and encoder block combination for ResNet-18 pre-trained on ImageNet. This observation implies that although feature encoders pre-trained on ImageNet have achieved good performance in MIL tasks in the literature \cite{lu2020data}, every new application of MIL requires searching for an optimal combination of feature extractor depth, decoder depth and MIL aggregator. It is also not clear if the performance achieved in MIL by ImageNet models is due to the sufficient discriminative features or due to the MIL aggregators exploiting higher order information at the WSI level.

Results for SS learning in Table \ref{tab:results} show that although SS learning was able to match the performance of the best ImageNet encoder and decoder combination, it was not able to match the performance of the MTL, MIC or MTL+MIC models. It is perhaps worth bearing in mind that SS models are known in practice to be hyper-parameter sensitive and usually need dataset specific design. Therefore the results we report may under-estimate the SS model performance. We would argue that supervised signal should be used when possible. 

Finally, following methods described in Desai \etal\cite{desai2020virtex}, we demonstrate attention plots for selected tokens in Figure \ref{fig:attention_plot}. In a colon tissue sample, the model appears to be focusing more on the main constituents of the colon tissue. For adipose (fatty) tissue sub-type, the model clearly focused on the middle, adipose rich area of the tissue slide. It is precisely these qualities that we believe provide some supporting evidence to our conjecture and explain the performance of the feature transfer from MIC or MTL+MIC to new tasks. Likewise, for functional qualities of a tissue image such as nerves, our model appears to identify correct areas where these are present. This allows both MIC and MTL+MIC models to perform well on the Atlas dataset \cite{hosseini2019atlas} whose labels identify functional traits of tissue in the image (dataset \#9 in Table \ref{tab:results}). 

The results above show the potential of using all available supervisory signals for pre-training general CP feature encoders. We also investigated the relative complexity of CP datasets and their computer vision counterparts within respective tasks. Intuitively, one can expect the intrinsic dimensionality of a dataset and a particular task to correlate well with the ability to learn general features from the dataset. Using intrinsic dimensionality estimation method described in Section \ref{sec:intrinsic-dim}, we compared datasets \#1, 2, 3, 4, 5, 6 and 7 with CIFAR100 and CIFAR10 \cite{krizhevsky2009learning} for classification; nuclei \cite{gamper2019pannuke}, epithelium \cite{janowczyk2016deep} nerve and vessel \cite{fraz2019fabnet} and gland \cite{awan2017glandular} segmentation datasets with CityScape \cite{Cordts2016Cityscapes}; and ARCH with MS-COCO for image captioning. We use the inverse of the area under the curve for a given dataset as a proxy to the relative intrinsic dimensionality of the dataset, shown  in Figure \ref{fig:intrinsic_dimensionality}. For both classification and segmentation tasks, we find that CP datasets are relatively less complex than their computer vision counterparts, particularly in the case of segmentation. ARCH, on the other hand, closely matches the complexity of MS-COCO -- we suspect that although ARCH is relatively smaller than MS-COCO, the size of the captions (see Figure \ref{fig:caption_lengths}) and the density of the information within them along with multiple instance nature of the dataset makes the learning process harder. 

\section{Limitations \& Future Directions}

\textbf{Experiments are limited to patch-level performance.} While we demonstrate the advantage of pre-training MIC or MTL+MIC models for transferring to completely unseen tasks, the real testing ground would be to evaluate the trained feature extractors on MIL or NIC tasks at the WSI level. Nonetheless, we are confident that there is a significant advantage in MTL+MIC pre-training as compared to transfering features from ImageNet in: ease of use - no need to optimise over network depth - and likely higher performance. While most of the datasets used in large-scale MIL studies are not publicly available, by making ARCH and used datasets publicly available in a single, easy to use repository we hope that the community will make the best use of the trained models. 

\textbf{Dataset construction leads to under-utilisation of transformer based architecture.} This work is a result of a laborious effort to extract and clean figures and their captions. While ARCH captions are dense in information, a large majority of the information in clinical papers and textbooks is discarded when constructing ARCH. Dosovitskiy \etal\cite{dosovitskiy2020image} have demonstrated that one could employ transformer based models only for image recognition, this opens the path to efficiently learn from semi-structured data such as clinical papers and textbooks with far less effort and our work clearly demonstrates the potential of such approach.

\textbf{Explosion in hyper-parameter settings.} One of the reasons for obtaining an encoder that provides general pathology image features is to avoid grid-search over the ImageNet encoder layer vs decoder depth combinations frequently seen in the literature, and consequently achieve wider adoption in practice. However, MTL, MIC or even SS like models are highly sensitive to hyper-parameters. On the one hand, this opens a call for stricter evaluation protocols; on the other hand, it is an opportunity for a unified pre-training procedure for CP that could utilise labelled patch-level, pixel-level and ARCH like dataset, as well as weakly labelled datasets such as TCGA \cite{weinstein2013cancer} and GTEx \cite{carithers2015novel}. CP community could draw inspiration from success of using Transformers \cite{vaswani2017attention} for pre-training in NLP that was able to achieve significant milestones for a seemingly simple pre-training, and yet provide empirical performance guarantees with respect to model size and dataset size \cite{kaplan2020scaling}.  

\section{Concluding Remarks}

In this work, we presented ARCH -- the largest multiple instance captioning dataset for pathology images to date -- carefully curated from textbooks and articles. We provided evidence to support our conjecture that pre-training on dense image captions in ARCH learns transferable representations for most CP tasks. It was shown that ARCH dataset is of similar complexity to its computer vision counterpart. We demonstrated the potential of new, yet unexplored sources of dense supervision in medical imaging, a field that is limited by the availability of high quality annotations. We posit that a model pre-trained on ARCH could be used without fine-tuning in multi-center screening, image retrieval, multiple instance or neural compression tasks and perform better than commonly used encoders trained on ImageNet. We are releasing ARCH along with the datasets and models used in this work in a single easy to use repository, such that the community could build on our work and fully explore the pre-training schemes that combine all of the available CP data.


{\small
\bibliographystyle{ieee}
\bibliography{egbib}
}

\end{document}